\documentclass{article}

\usepackage{microtype}
\usepackage{graphicx}
\usepackage{booktabs} 

\usepackage[accepted]{icml2020}

\icmltitlerunning{Scene Graph Reasoning for Visual Question Answering}

\usepackage{graphicx}
\usepackage{subcaption}
\usepackage[export]{adjustbox}
\usepackage{amsmath, amssymb}
\usepackage{dsfont}

\begin{document}

\twocolumn[
\icmltitle{Scenes and Surroundings: Scene Graph Generation using Relation Transformer}




\begin{icmlauthorlist}
\icmlauthor{Rajat Koner}{lmu}
\icmlauthor{Poulami Sinhamahapatra}{frh}
\icmlauthor{Volker Tresp}{siemens,lmu}
\end{icmlauthorlist}

\icmlaffiliation{lmu}{Ludwig Maximilian University of Munich, Munich, Germany}
\icmlaffiliation{siemens}{Siemens AG, Munich, Germany}
\icmlaffiliation{frh}{Fraunhofer IKS, Munich, Germany}

\icmlcorrespondingauthor{Rajat Koner}{koner@dbs.ifi.lmu.de}

\icmlkeywords{Visual question answering, Scene graphs, Graph learning, Reinforcement learning}

\vskip 0.3in
]



\printAffiliationsAndNotice{} 


\begin{abstract}
Identifying objects in an image and their mutual relationships as a scene graph leads to a deep understanding of image content. Despite the recent advancement in deep learning, the detection and labeling of visual object relationships remain a challenging task. This work proposes a novel local-context aware architecture named relation transformer, which exploits complex global objects to object and object to edge (relation) interactions. Our hierarchical multi-head attention-based approach efficiently captures contextual dependencies between objects and predicts their relationships. In comparison to state-of-the-art approaches, we have achieved an overall mean \textbf{4.85\%} improvement and a new benchmark across all the scene graph generation tasks on the Visual Genome dataset.
\end{abstract}
\section{Introduction}
\label{sec:introduction}
\begin{figure}[ht]
\begin{subfigure}{0.5\textwidth}
\centering
\includegraphics[width=0.7\linewidth]{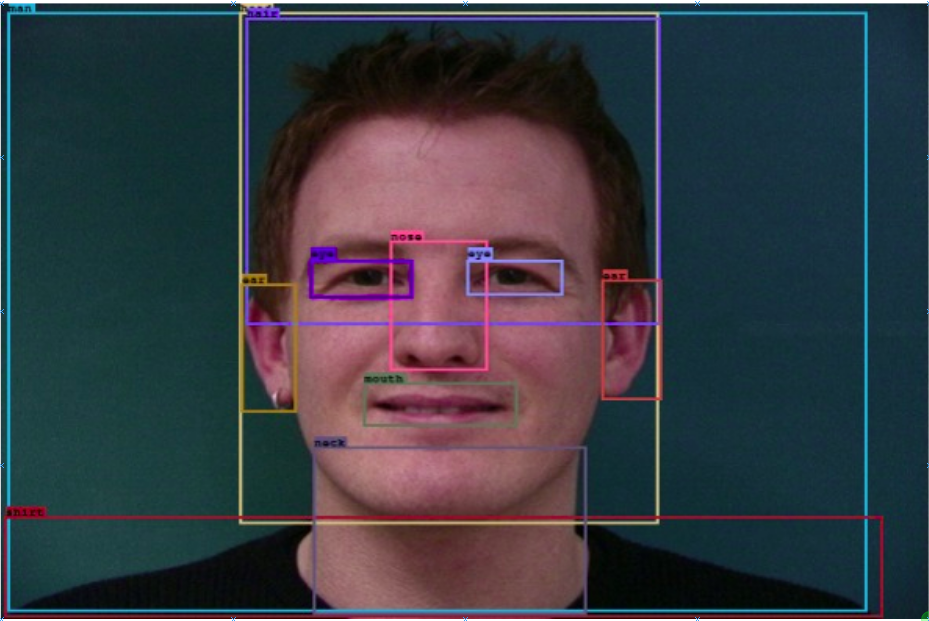} 
\caption{Scene consisting of a man's  face}
\label{fig:subim1}
\end{subfigure}
\begin{subfigure}{0.5\textwidth}
\centering
\includegraphics[width=0.7\linewidth, height=4cm]{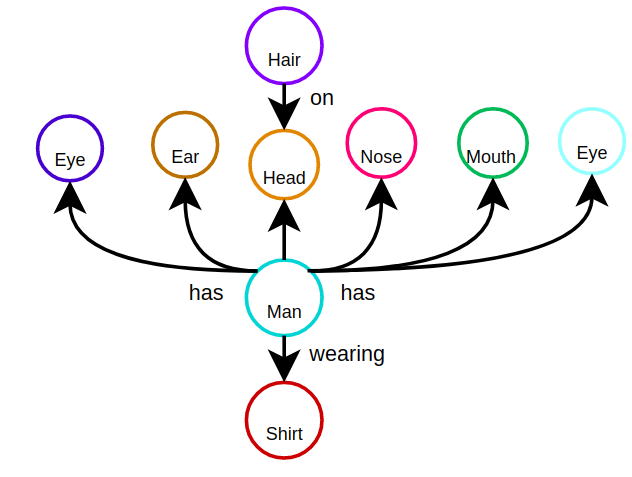}
\caption{Corresponding scene graph}
\label{fig:subim2}
\end{subfigure}
\caption{\ref{fig:subim1} is an example image  of a face of a man. \ref{fig:subim2} describes the corresponding  scene graph,  annotated with various objects like head, ear, shirt (color coded as the respective bounding box) and their mutual relationships.}
\label{fig:man_sg}
\end{figure}

A \textit{scene graph} is a graphical representation of an image consisting of multiple entities and their relationships expressed in a triple format like $\langle subject, predicate, object\rangle$. Objects in the scene become \textit{nodes} in the graph, and a directed \textit{edge} denotes a mutual relationship or predicate. 
Fig. \ref{fig:man_sg}, \lq Eye\rq,\lq Hair\rq,\lq Head\rq,\lq Man\rq\: are objects or nodes and their mutual relationships are described by the predicates 
 \lq has\rq,\lq on\rq.  
 
 Automated scene graph generation is executed in  two steps:
 first, detection of the objects present in an image. Second, predicates among objects are derived. Current state-of-the-art object detection approaches have achieved impressive performance in spatially locating objects, while those for relation prediction are still in a nascent stage. 
 To achieve state-of-the-art performance, it is important to consider context information (which could be both local or global) to model dependencies between objects and predicates. An  extracted scene graph can be used in many applications like visual question answering\cite{ghosh2019generating,hildebrandt2020scene}, image retrieval\cite{schuster2015generating}, image captioning\cite{li2017scene}. 
 
 The primary challenge involved in scene graph generation is to understand the role of each object in an image and how objects are related or influenced by others in the context of the whole scene. For example,  in Fig. \ref{fig:man_sg}, the presence of nodes like \lq Eye \rq, \lq Hair\rq, \lq Nose\rq, \lq Head\rq, indicate that these together describe a face. 
 Additionally, node \lq Shirt\rq\: implies that this is a face of a  \lq Human\rq and not an animal. 
Node dependencies are also crucial for predicting an edge or a  pairwise relation. 
 Conversely, spatial and semantic co-occurrence
 also helps in identifying node classes.
 A subsequent challenge is to predict correct predicates describing the exact relationship between two objects.
 
In this paper, we propose a novel scene graph generation architecture named \textbf{Relation Transformer}, which leverages upon interactions among objects, predicates, their respective influence on each other, as well as their co-occurrence pattern. 
Based on the above mention challenge, we have modified the transformer architecture with some novel changes. To summarize our contributions:

\begin{itemize}
\setlength\itemsep{0pt}
\item We introduce a novel positional encoding algorithm for edges in the transformer decoder that accumulates global scene context while preserving local context. This is specifically useful since an edge label can often be predicted from the head or tail entity class. 


\item An algorithm that predicts an edge label needs to be aware of all node labels of other entities present in the scene and other edge labels. We have applied unrestricted attention and custom ordering of the E2N and E2E\footnotemark[0] blocks to achieve this.
\item We have achieved an overall mean 4.85\%  improvement overall scene graph generation tasks and set a new benchmark on the Visual Genome dataset.
\end{itemize}
\footnotetext[0]{According to the context in the paper, we named encoder self-attention to N2N(Node to Node), decoder cross-attention to E2N(Edge to Node), and decoder self-attention to E2E(Edge to Edge) attention.}

\begin{figure*}
\includegraphics[width=\textwidth,height=5cm]{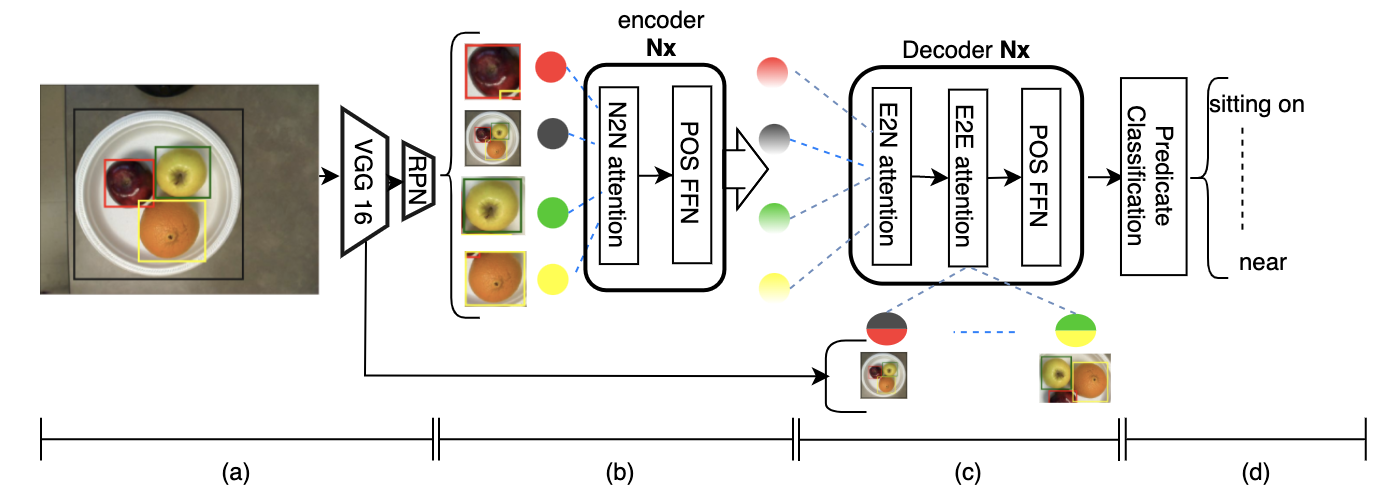}
 \caption{An overview of the proposed Relation Transformer architecture. The network consists of four stages: a) Feature generation by an object detector and bounding box extraction using RPN, b) Creation of context-rich node embeddings (light color) using N2N attention from initial 
 nodes (dark color) c) Creation of  edge embedding (bicolor based on respective nodes)  using context from all nodes (E2N) and then from other edges (E2E), d) Classification of  the relation using $\langle subject, edge, objects \rangle$ manner. Best viewed in color.}
\label{fig:atten4rel}

\end{figure*}

\if false
\subsection{Related Work}

Various models have been proposed that perform VQA on both real-world and artificial datasets. Currently, most dominant VQA approaches can be categorized into two different branches: First, monolithic neural networks which perform implicit reasoning on latent representations obtained from fusing the two data modalities. Second,  multi-hop methods that form explicit symbolic reasoning chains on a structured representation of the data. Monolithic network architectures obtain visual features from the image either in from individual detected objects or by processing the whole image directly via convolutional neural networks (CNNs). The derived embeddings are usually scored against a fixed answer set along with the embedding of the question obtained from a sequence model. Moreover, co-attention mechanism are frequently employed to couple the vision and the language models allowing for interactions between objects from both modalities \cite{kim2018bilinear, anderson2018bottom, cadene2019murel, yu2017multi, zhu2017structured}. Monolithic networks are among the dominant methods on previous real-world VQA datasets such as \cite{antol2015vqa}. 
However, they suffer from the black-box problem and posses limited reasoning capabilities with respect to complex questions that require long reasoning chains (see \cite{chen2019meta} for a detailed discussion).

Explicit reasoning methods combine the sub-symbolic representation learning paradigm with symbolic reasoning approaches over structured representations of the image. Most of the popular explicit reasoning approaches follow the idea of neural module networks (NMNs) \cite{andreas2016neural} which perform a sequence of reasoning steps realized by forward passes through specialized neural networks that each correspond to predefined reasoning subtasks. Thereby, NMNs construct functional programs by dynamically assembling the modules resulting in a question-specific neural network architecture. In contrast to the monolithic neural network architectures described above, these method contain a natural transparency mechanism via the functional programs. However, while NMN-related methods (e.g., \cite{hu2017learning, mao2019neuro}) exhibit good performance on synthetic datasets such as CLEVR \cite{johnson2017clevr}, they require functional module layouts as additional supervision signal to obtain good results. Hence, NMNs have a limited applicability in most real-world settings. In this paper, we propose a novel neuro-symbolic scene graph reasoning approach which leverages representation learning combined with explicit reasoning on scene graphs. Closely related to our method is the Neural State Machine (NSM) proposed by \cite{hudson2018compositional}. The underlying idea of NMS consists in first constructing a scene graph from an image and treat it as a state machine. Concretely, the nodes correspond to states and edges to transitions. Then conditioned on the question, a sequence of instructions is derived that indicates how to traverse the scene graph and arrive at the answer. In contrast to NSM we treat path-finding as a decision problem in a reinforcement learning setting. Concretely, we outline in the next section how extracting predictive paths from scene graphs can be naturally formulated in term of a goal-oriented random walk induced by a stochastic policy that allows to balance between exploration and exploitation. Moreover, our framework integrates state-of-the-art techniques from graph representation learning and NLP. In this paper we only consider basic policy gradient methods but more sophisticated reinforcement learning technique lo s can be be employed in future works.


\paragraph{Statistical Relational Learning} Machine learning methods for KG reasoning aim at exploiting statistical regularities in observed connectivity patterns. These methods are studied under the umbrella of statistical relational learning (SRL) \cite{nickel2015review}. In the recent years KG embeddings have become the dominant approach in SRL. The underlying idea is that graph features that explain the connectivity pattern of KGs can be encoded in low-dimensional vector spaces. In the embedding spaces  the interactions among the embeddings for entities and relations can be efficiently modelled to produce scores that predict the validity of a triple. Despite achieving good results in KG reasoning tasks, most embedding based methods hardly capture the compositionality expressed by long reasoning chains. 

Recently, multi-hop reasoning methods such as MINERVA  \cite{minerva} and DeepPath \cite{wenhan_emnlp2017} were proposed. Both methods are based on the idea that a reinforcement learning agent is trained to perform a policy-guided random walk until the answer entity to a query is reached. Thereby, the path finding problem of the agent can be modeled in terms of a sequential decision making task framed as a Markov decision process (MDP). The method that we propose is in this work follow a similar philosophy in the sense that we train an RL agent to navigate on a scene graph to the correct answer node. However, a conceptial difference is that he agents in MINERVA and DeepPath perform walks on large scale knowledge graphs exploiting repeating, statistical patterns. Thereby, the policies implicitly incorporate (fuzzy) rules. In addition, instead of processing free-form questions, the query in the KG reasoning setting is structured as a pair of symbolic entities. That is why we propose a wide range of modification to adjust our method to the challenging VQA setting.

\subsection{Connection to Cognitive Science}
VOLKER

\fi
\section{Method}
\label{sec:our_method}
We formulate scene graph generation task as a multi-hop attention based context propagation problem between nodes and edges. This task is decomposed into four sub-tasks, 
starting with object detection, 
followed by modelling interactions between the nodes, 
then accumulating influence from both nodes and edges, 
and, finally, classifying relations between the objects. Below we will describe these sub-tasks, along with a brief introduction of the attention mechanism, the transformer, and their roles in these modules. An overview of the proposed Relation Transformer architecture is shown in Fig. \ref{fig:atten4rel}.

\subsection{Problem Decomposition}
A scene graph $G = (N, E)$ of an image $I$  is used for describing each node or object ($n_i\in N$) and their interlinked relations (like geometric, spatial etc.) with a directed edge ($e_{ij}\in E$). A set of nodes $\{n_i\}$, can be represented by their corresponding bounding boxes as B = \{$b_1,b_2,..b_n$\}, $b_i\in \mathbb{R}^4$ and their class label O = \{$o_1,o_2..o_n$\}, $o_i\in C$. Each relation $r_{{sub}\rightarrow {obj}}\in R$ defines the relationship between the subject and object node. Hence, scene graph generation can be formulated as  a three factor model as,
\begin{align}
\small
Pr(G \rvert I) = Pr(B \rvert I)\: Pr(O \rvert B,I)\: Pr(R \rvert O,B,I). 
\end{align}
$Pr(B \rvert I)$ can be inferred by any object detection model (Sec. \ref{od}). Sec.\ref{cpo} describes conditional probability of an object class $Pr(O \rvert B,I)$, where the presence of one object can be influenced by another class. To model the relationships $Pr(R \rvert O,B,I)$, we first compute an undirected edge
(Sec. \ref{cee}) between two objects, then conclude on a directed edge($r_{{sub}\rightarrow {obj}}$) (Sec. \ref{rc}).

\subsection{Object Detection}\label{od}
We have used Faster-RCNN  \cite{ren2015faster} with a VGG-16  \cite{simonyan2014very} backbone for object detection. 
For $i^{th}$ object candidate, 
we obtain
visual features $v_i^{\textit{RoI}} \in \mathbb{R}^{4096}$, 
bounding box coordinates $b_i \in \mathbb{R}^5$ 
and class label probabilities\footnote{GloVe embeddings for all classes has been used with a dimension of 200.} $ o^{\textit{init}}_i \in \mathbb{R}^{200} $. The initial feature ($n_i^{\textit{in}} \in \mathbb{R}^{2048}$) of $i^{th}$ node is obtained by applying a linear projection layer($f_{\textit{nlp}}$) on its concatenated features as described in Eq. \ref{nin}. We have considered these individual proposals and their respective features as the initial node embeddings of the scene graph.
\begin{equation}\label{nin}
n^{\textit{in}}_i = f_{\textit{nlp}}([v^{\textit{RoI}}_i, o^{\textit{init}}_i, b_i])
\end{equation}

\subsection{Context Propagation:}\label{cp}
The core idea of our approach is the efficient context propagation across all nodes and edges using a transformer encoder-decoder architecture  \cite{vaswani2017attention}. At the heart of the transformer lies a self-attention mechanism, which is briefly described next.

\subsubsection{Attention:}\label{at}
Attention mechanisms enable multi-hop  information propagation in sequences ans graphs. The transformer   \cite{vaswani2017attention} architecture uses self-attention mechanisms for mapping of the global dependencies. One defines attention as:
\vspace{-0.5cm}
\begin{align}\label{attention}
    \text{Attention}(Q,K,V) = \text{softmax}(\dfrac{QK^T}{\sqrt{d_k}})V .
\end{align}
The last equation describes a self-attention function, where query(Q), keys(K), and values(V) are a set of learnable matrices, and $d_k$ is the scaling factor. The output is computed as a weighted sum of the values, where the weight assigned to each value is computed by multiplying a query matrix with its corresponding key. 

\subsubsection{Context Propagation for Objects:}\label{cpo}
Contextualization of objects not only enhances object detection   \cite{liu2018structure} by exploring the surroundings of objects, but also encodes more expressive features for relation classification. For $i^{th}$
node, we have used initial features from Eq. \ref{nin} along with a positional ($\textit{pos\_enc}^{n}\in \mathbb{R}^{2048}$) feature vectors, based on the actual position of the node in the sequence, 
\begin{align}\label{nfinal}
\small
n^{\textit{final}}_i = \text{encoder}(n^{\textit{in}}_i+\textit{pos\_enc}^{n}(n^{\textit{in}}_i)).\\
\small
o^{\textit{final}}_i = \text{argmax}(f_{\textit{classifier}}(n^{\textit{final}}_i)).\label{ofinal}
\end{align}
After contextualization of the nodes by the encoder\footnote{our encoder block remains same as Transformer, and its architecture shown in Figure \ref{fig:atten4rel}.} in Eq. \ref{nfinal}, we have obtained final node features ($n^{\textit{final}}_i$). Final node features are subsequently used for two purposes.  Firstly,  they are passed to a linear object classifier (Eq. \ref{ofinal}) to get the final object class ($o^{\textit{final}}_i \in C$) probability and finally, the same node features are passed to the next module for edge context propagation.

\begin{table*}[t!]
\centering

\begin{tabular}{lccc|ccc|cc|cc|c}
\toprule
\multirow{2}{*}{ \textbf{\phantom{abcdef}Model} }    
& \multicolumn{6}{c}{Graph constraint}  
& \multicolumn{4}{c}{No graph constraint} 
& \multirow{3}{*}{ \textbf{Mean}}  \\
                                &\multicolumn{3}{c}{\textbf{SGCLS}} &\multicolumn{3}{c}{\textbf{PRDCLS}}
                                &\multicolumn{2}{c}{\textbf{SGCLS}} &\multicolumn{2}{c}{\textbf{PRDCLS}}  \\
\textbf{\phantom{abcd}Recall@}                      & 20                   & 50                   & 100                                                               & 20                   & 50                   & 100                                                     & 50 &100  & 50 & 100      &                    \\ 
\midrule
Message Passing   \cite{Xu_2017_CVPR}       & 31.7  & 34.6   & 35.4                     & 52.7  & 59.3  & 61.3                      &  43.4  & 47.2                           
&  75.2  &  83.6   
& 52.44\\
Associative Embedding   \cite{Xu_2017_CVPR} & 18.2  & 21.8  & 22.6                      & 47.9  & 54.1  & 55.4 
& 26.5 & 30.0    
& 68.0  &  75.2  
&41.17\\
MotifNet(Left to Right)   \cite{zellers2018neural}              
& 32.9  & 35.8  & 36.5                      & 58.5  & 65.2  & 67.1                      & 44.5    & 47.7   
&81.1   & 88.3   
&55.76\\
Large Scale VRU  \cite{zhang2019large}      & 36.0  & 36.7  & 36.7                      & 66.8  & 68.4  & 68.4                      &  -  &  -   
&  -  &  -   
&52.16\\
ReIDN   \cite{zhang2019graphical}           & 36.1  & 36.8  & 36.8                      & 66.9  & 68.4  & 68.4                      &  48.9  & 50.8                          
&  93.8  &  97.8   
&60.49\\ 
\midrule
{\bfseries}\textbf{Relation Transformer (Ours)}  
& \textbf{43.4}  & \textbf{43.6} & {\textbf{43.7} } & \textbf{68.1}        
& \textbf{68.5}        & \textbf{68.5}  
& \textbf{60.6}   & \textbf{61.7}
& \textbf{96.5}    & \textbf{98.8}          & \textbf{65.34}\\ 
\bottomrule
\end{tabular}
\caption{Comparison of our model with state of the art methods tested in Visual Genome \cite{krishna2017visual}}
\label{tab:results}
\vspace{-0.5cm}
\end{table*}

\subsubsection{Context Propagation for Edges}\label{cee}
In this module, edge features are captured by accumulating context information across all nodes and edges. Edges are highly dependent on the local context, as they are associated with only a pair of nodes (subject, object). 
 We have introduced novel changes in decoder, such that the network learns relational(E.g. spatial, semantic) influences from other nodes or edges by exploiting both local and global contexts.

For an edge belonging to $i^{th}$ and $j^{th}$node, visual features $e_{ij}^{\textit{vis}} \in \mathbb{R}^{4096}$ consist of the union of two object boxes $b_{i,j}$ as shown in Figure \ref{fig:atten4rel}. 
Afterwards, spatial features $b_{i,j}\in \mathbb{R}^5$($b_i$ and $b_j$) are added with the concatenated GloVe  \cite{pennington2014glove} embedding ($e^{\textit{sem}}_{ij}$) of both classes.
Subsequently, a linear projection layer ($f_{\textit{elp}}$) is used to obtain the initial edge embeddings ($e^{\textit{in}}_{i,j} \in \mathbb{R}^{2048}$) as 
\begin{align}\label{ein}
\small
 e^{in}_{i,j} = f_{elp}(e_{ij}^{vis}+b_{ij}+e^{sem}_{ij})
\end{align}

As mentioned earlier, we have introduced three modifications in the transformer decoder network such that it models the interaction between nodes and edges efficiently.
\begin{enumerate}
\setlength\itemsep{0pt}
\item The decoder masked attention has been removed so that it can attend to the whole sequence, not just part of it.
\item A novel positional encoding vector has been introduced ($\textit{pos\_enc}^{\textit{e}_{ij}}\in \mathbb{R}^{2048}$) for edges ($e^{\textit{in}}_{i,j}$) that encodes the position of both the source nodes, instead of the position of the edge alone. We hypothesize that it will be beneficial for the network to distinguish the source nodes (subject and object) out of all distinct nodes and the corresponding edges between source nodes to the other edges. This design bias can accumulate the global context without losing its focus on the local context or source nodes.
\begin{equation} \label{pee}
\begin{split}
\small
\textit{pos\_enc}^{e_{ij}}_{(k,k+1)} &=[\sin(p_i/m^{2k/d_{dim}}),\cos(p_i/m^{2k/d_{dim}})]. \\
\small
\textit{pos\_enc}^{e_{ij}}_{(k+2,k+3)} &=[\sin(p_j/m^{2k/d_{dim}}),\cos(p_j/m^{2k/d_{dim}})].
\end{split}
\end{equation}
Eq. \ref{pee} describes positional encoding for an edge, where $p_i$ and $p_j$ are the positions of the nodes $n_i$ and $n_j$, $m$ is maximum number of sequence,  $d_{\textit{dim}} \in \mathbb{R}^{2048}$ is same dimension as $e^{\textit{in}}_{i,j}$, and k denotes the $k^{th}$ position in the positional encoding features vector.

\item The order of self-attention applied in the decoder has been altered. At first, E2N self-attention has been applied from an edge to all the nodes. Then, E2E attention from an edge to all the edges has been incorporated. Since, the edge is created from only two nodes, E2N attention accumulates necessary global context from all nodes. Afterwards, E2E attention will help an edge, enriched with global context, to learn from edges with similar relational embedding. Finally, we get contextual edge features ( $e^{\textit{final}}_{i,j} \in \mathbb{R}^{2048}$)  as, 
\begin{align}\label{efin}
\small
e^{\textit{final}}_{i,j} =\text{decoder}(e^{\textit{in}}_{i,j} + {pos\_enc}^{e_{ij}})
\end{align}
\end{enumerate}

\subsection{Relation Classification}\label{rc}
 A  relation is a directional property, i.e., \textit{subject} and \textit{object} cannot be exchanged.  After obtaining the context-riched node and edge embeddings, a joint relational embedding ($rel_{\textit{emb}} \in \mathbb{R}^{2048}$) has been created consisting of triplets like $\langle subject, edge, object \rangle$ followed by a Leaky ReLU  \cite{xu2015empirical} non linearity for the predicate classification as described in Eq. \ref{rel}. Finally, to get the \textit{softmax} distribution of a  predicate a fully connected layer ($W_{\textit{final}}$) along with the Frequency Baseline \cite{zellers2018neural} has been added to model as described in Eq. \ref{rfin}.
\begin{align}\label{rel}
\small
\setlength\itemsep{-0.5pt}
rel_{\textit{emb}} = \text{LReLU}(f_{\textit{rel}}([n^{\textit{final}}_i,e^{\textit{final}}_{i,j},n^{\textit{final}}_j]))
\end{align}
\begin{align}
\begin{split}
\small
\setlength\itemsep{-0.5pt}
Pr(R \rvert B,O,I ) = \text{softmax}(W_{\textit{final}}(rel_{\textit{emb}})+fq(\textit{sub},\textit{obj})) \label{rfin}
\end{split} 
\end{align}

\section{Experiments}
\label{sec:experiments}

\paragraph{Dataset and Experimental Setup}
We used  Visual Genome(VG) \cite{krishna2017visual} for our training and evaluation.  It is one of the largest and most challenging dataset on scene graph generation for real world images. To have a fair comparison with present state-of-the-art models \cite{zellers2018neural,newell2017pixels,zhang2019graphical,zhang2019large}, we have used the same refined version of VG proposed in \cite{Xu_2017_CVPR} along with their official split. This dataset contains the most frequently occurring 150 objects and 50 relationships of VG. We have followed the same evaluation as in the current benchmark \cite{zhang2019graphical} and computed scene graph classification (SGCLS) and predicate classification (PREDCLS). 

\paragraph{Results and Discussion}
Table \ref{tab:results}, shows the performance of our method in comparison with other methods. Here, methods such as \cite{Xu_2017_CVPR,zellers2018neural} have used various techniques of context propagation, while ReIDN\cite{zhang2019graphical} and VRU \cite{zhang2019large} have used special losses (E.g. contrastive loss) for better modelling of scene graph. Table \ref{tab:results} demonstrates that our novel context propagation for both objects and edges significantly improves the performance even with simple cross-entropy loss.

Additionally, analysis of false prediction provide a great insight that the network learned semantically plausible closer outputs. E.g. `on' is the most mispredicted relation in evaluation settings, which is 56.9\% times predicted as `of' for (sub., obj.) like (face, woman), (wing, plane). Interestingly, the mispredicted `face of woman' is more appropriate than `face on woman', indicating a network not necessarily failing to predict correctly rather due to a huge bias in the dataset to 34.3\% `on' relations. 
More such positive and negative examples have been listed out in Supplementary. 
Also, as mentioned in ``No Graph Constraint", the high recall in PREDCLS (98.8\%) indicates even if the network failed to predict the actual relation in top most prediction, it is mostly captured when multiple relations are being allowed in the subject-object pair.
\section{Conclusion}
\label{sec:conclusion}
We have proposed a novel method for visual question answering based on multi-hop sequential reasoning and deep reinforcement learning. Concretely, an agent is trained to extract conclusive reasoning paths from scene graphs. 
To analyze the reasoning abilities of our method in a controlled setting, we conducted a preliminary experimental study on manually curated scene graphs and concluded that our method reaches human performance. In future works, we plan to  incorporate state-of-the-art scene graph generation into our method to cover the complete VQA pipeline.

\newpage
\bibliography{main}
\bibliographystyle{icml2020}
\newpage

\begin{center}
  {\bfseries Supplementary Material}\\
\end{center}
This is a supplementary material for our paper \lq Relation Transformer Network\rq. Here, we will discuss more about implementation details, attention map and qualitative results conducted on Visual Genome dataset. More detailed work can be found in \cite{koner2020relation}.
\begin{itemize}
\item{\textbf{Implementation Details:}}
In this section we will list out hyper-parameter used in final model.
\begin{enumerate}
    \item optimizer : Stochastic Gradient Descent(SGD)
    \item learning rate : $10^{-3}$ with reduce on plateau and patience $3$
    \item batch size : 16
    \item dropout : $0.25$
    \item Context Propagation of Objects : 3 E2N modules
    \item Context Propagation for Edges : 2 E2N modules
    \item attention head : 12 attention heads are used in both N2N and E2N.
    \item Directed Relation Prediction Module (RPM): As discussed in paper, a RPM module leverages upon context rich nodes  (e.g. $n_i^{final},n_j^{final}$) and undirected edges ($e_{ij}^{final}$) to produce final directed relation embedding between two nodes ($rel_{i \rightarrow j}^{final}$). The input to RPM ($rel_{i \rightarrow j}^{in}$) is normalized by Layer Normalization then followed by a linear layer ($W_{1} \in \mathbb{R}^{4096} $), dropout then another linear layer ($W_{2} \in \mathbb{R}^{2048} $) and finally followed by Leaky ReLU non-linearity.
    
    \item random seed : 42
\end{enumerate}
\item{\textbf{Analysis of Attention:}}
Here, we present an  analysis of how attention mechanisms help in scene understanding.
In our approach, attention has been used for context propagation between nodes (N2N) and between edge to node (E2N). This interaction has been visualized using an attention heatmap in Fig. \ref{fig:attention map}. Here mutual influence between each pair or row and column is plotted using a score between 0 to 1, where 1 signifies maximum influence, 0 is for minimum. We have used attention mask from top most layer for both module. \\
In Fig. \ref{fig:attention map} (left), a scene with a seagull flying near the beach is shown. Its corresponding node to node (N2N) attention map exhibits detected objects like \lq bird\rq, \lq wing\rq, \lq tail\rq, \lq beach\rq \: and indicates which nodes or objects are more influential for joint object and relation classification. For example, the node \lq bird \rq \: has high attention for \lq bird\rq, \lq wing\rq, \lq tail\rq , that suggests what are the nodes related to it and what could be their potential relationships. Moreover, \lq wing\rq \: has high attention with \lq beach\rq \:  that could be a potential indicator of influence, suggesting relationship could be flying over the beach. This is further confirmed by  attention score for edge \lq beach-bird\rq\:in edge to node (E2N) attention. For other edges like \lq bird-tail \rq, \lq bird-wing\rq, \: \lq bird\rq \: could be the most influential node for these edges, thus provide a clear intuition about the kind of relationship that could exists among these nodes.\\
In Fig. \ref{fig:attention map} (middle), nodes like \lq man\rq, \lq trunk\rq, \lq ski\rq  \: and their mutual high attention score provide context interpretability. Also, its associated edge \lq man-ski\rq \: shows high influence for all nodes, that reflects context awareness of the edge.
Similarly, in Fig. \ref{fig:attention map} (right), the nodes like \lq glove\rq, \lq hair\rq, \lq hand\rq,
shows high mutual influence in node to node (N2N) attention heatmap. Also, \lq glove\rq and  \lq sink\rq \: show high attention indicating contextual influence. The relationships are further derived from edge to node (E2N) attention where edges like \lq glove-woman\rq, \lq hair-woman\rq, \lq glove-hand\rq \: show high attention with node \lq woman\rq \: 
suggesting that the scene consists of a woman who has hair 
and that the woman is wearing glove on her hand.\\

\begin{figure}[h!]
\centering
    \begin{subfigure}[b]{\linewidth}
        \centering
        \includegraphics[width=0.3\linewidth, height=3cm]{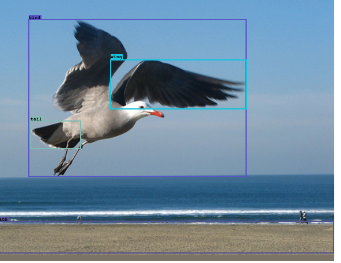}
        \hfill
        \includegraphics[width=0.3\linewidth, height=3cm]{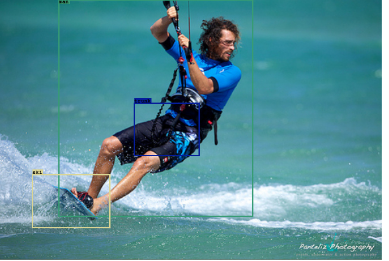}
        \hfill
        \includegraphics[width=0.3\linewidth, height=3cm]{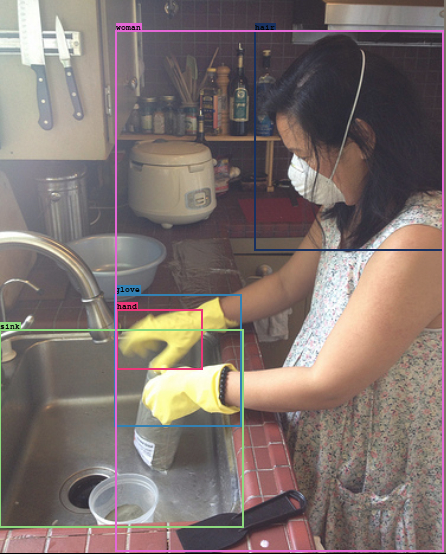}
        \caption{Scenes with objects and bounding boxes with respective detected labels}
    \end{subfigure}    
     \vskip\baselineskip
     \begin{subfigure}[b]{\linewidth}
        \centering
        \includegraphics[width=0.3\linewidth, keepaspectratio]{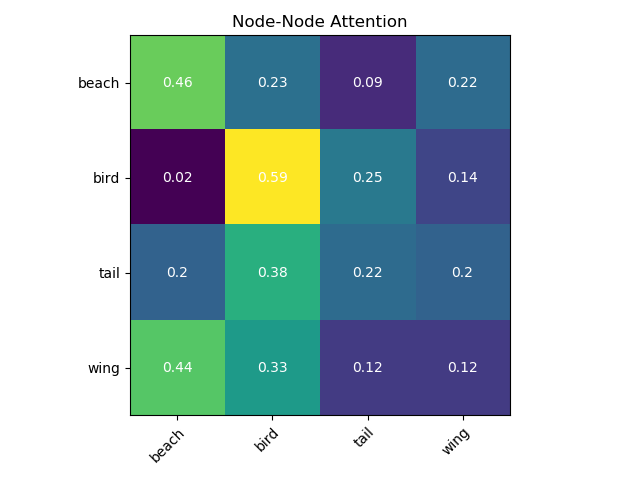}
        \hfill
        \includegraphics[width=0.3\linewidth, keepaspectratio]{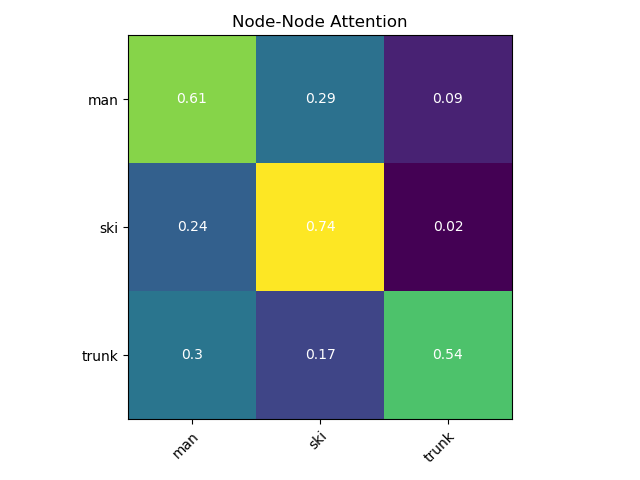}
        \hfill
        \includegraphics[width=0.3\linewidth, keepaspectratio]{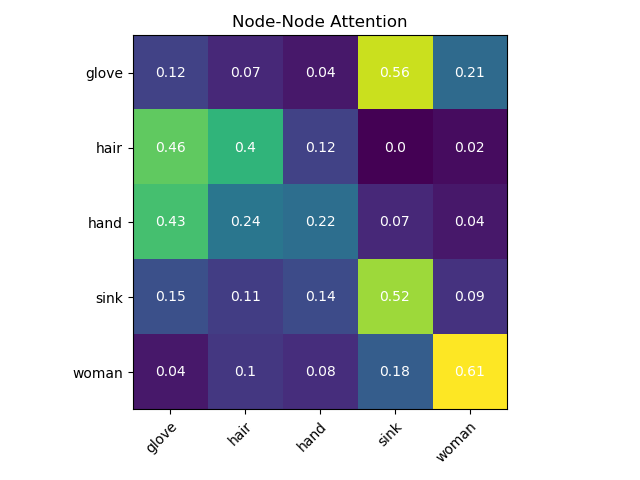}
        \caption{Node to Node Attention heatmap}
     \end{subfigure}
     
   \vskip\baselineskip
     \begin{subfigure}[b]{\linewidth}
        \centering
        \includegraphics[width=0.3\linewidth]{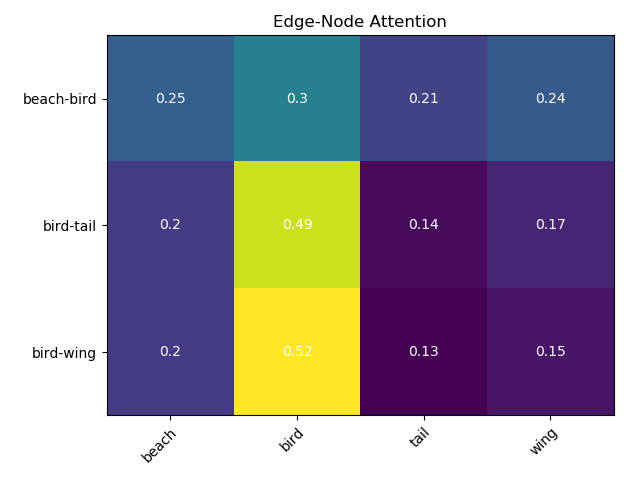}
        \hfill
        \includegraphics[width=0.3\linewidth]{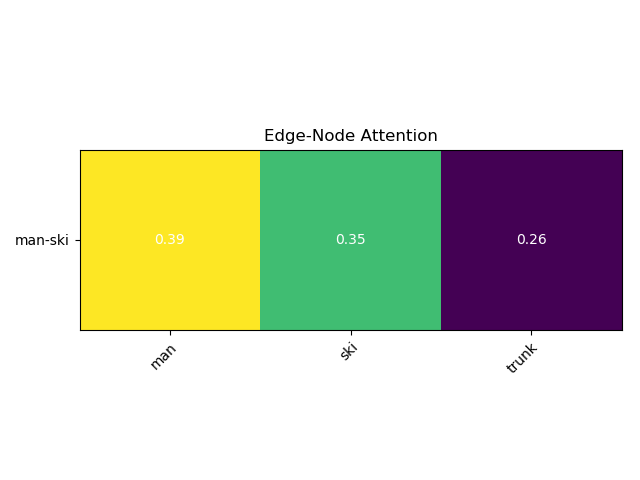}
        \hfill
        \includegraphics[width=0.3\linewidth]{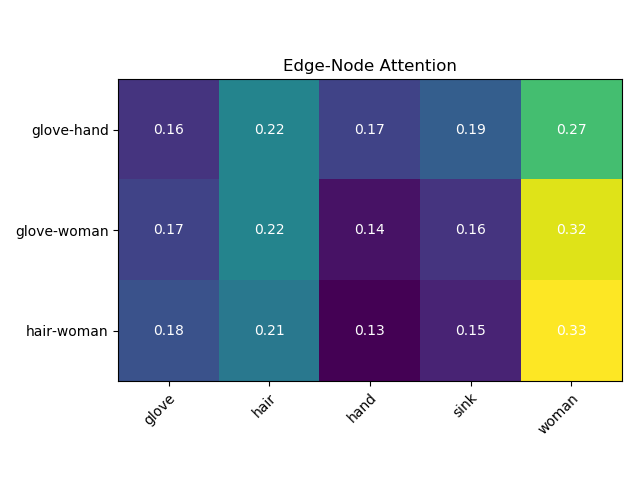}
        \caption{Edge to Node Attention heatmap}
     \end{subfigure}


    \vskip\baselineskip
     \begin{subfigure}[b]{\linewidth}
        \centering
        \includegraphics[width=0.27\linewidth]{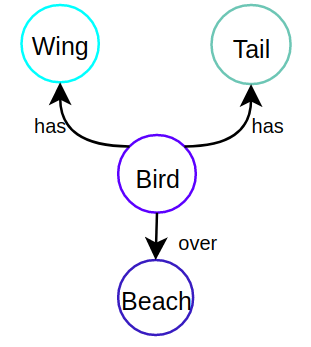}
        \hfill
        \includegraphics[width=0.13\linewidth]{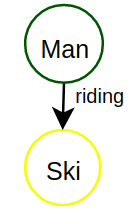}
        \hfill
        \includegraphics[width=0.3\linewidth]{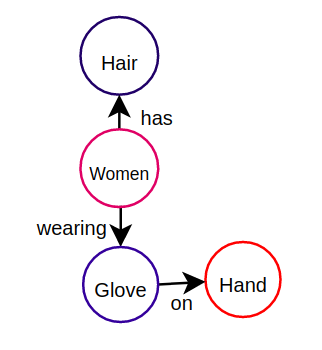}
        \caption{Generated scene graph}
     \end{subfigure}

\caption{Some example output from our network with associated attention map and scene graph.}
\label{fig:attention map}
\end{figure}

\item \textbf{Qualitative Results:} In this section, we will provide a few more qualitative samples generated by our network in both positive and negative scenarios. To improve visibility and interpretability, we only consider the interaction among ground truth objects and relations in these examples.\\

Fig. \ref{fig:attention map positive} (left column), shows the positive scenario, where our network is able to detect correct relationships label despite the presence of repetitive bounding box (boy and child) or similar objects (giraffe). Thus, it shows the robustness of our method.\\

Fig. \ref{fig:attention map negative} (right column), shows the negative scenario, where network prediction is different from ground truth labels. In most of these cases, it was found that predicted labels are semantically closer to ground truth labels, and from a human perspective, both could be right. For example \textit{man-at-beach} and \textit{man-on-beach} both are grammatically correct. As discussed in the paper, one of the reason for this is biases present in the training dataset .
\begin{figure}[h]
\centering
    \begin{subfigure}[b]{\linewidth}
        \centering
        \includegraphics[width=0.45\linewidth, height=4cm]{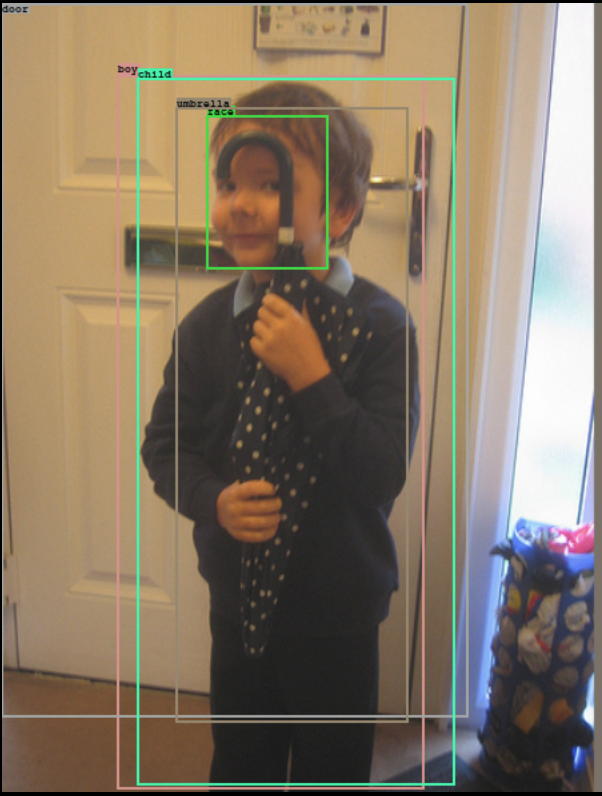}
        \hfill
        \includegraphics[width=0.45\linewidth, height=4cm]{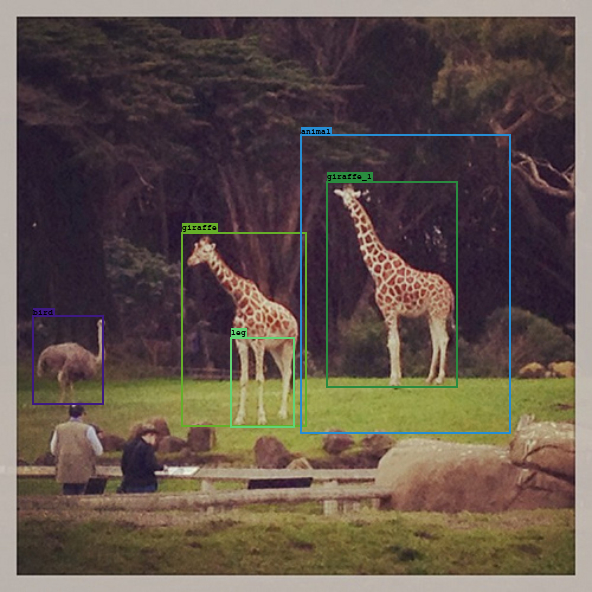}
        \caption{Scenes with objects and bounding boxes with respective labels}
    \end{subfigure}    
     \vskip\baselineskip
     \begin{subfigure}[b]{\linewidth}
        \centering
        \includegraphics[width=0.45\linewidth, keepaspectratio]{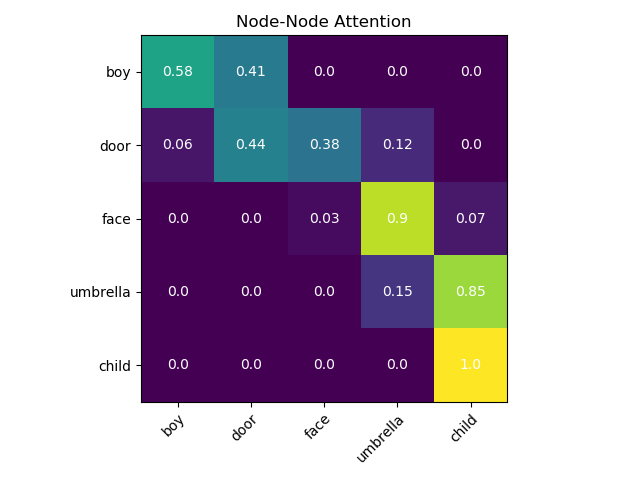}
        \hfill
        \includegraphics[width=0.45\linewidth, keepaspectratio]{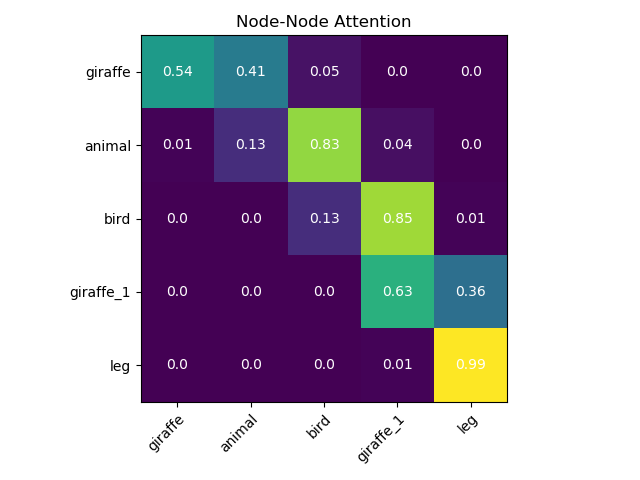}
        
        \caption{Node to Node Attention heatmap}
     \end{subfigure}
     
   \vskip\baselineskip
     \begin{subfigure}[b]{\linewidth}
        \centering
        \includegraphics[width=0.45\linewidth]{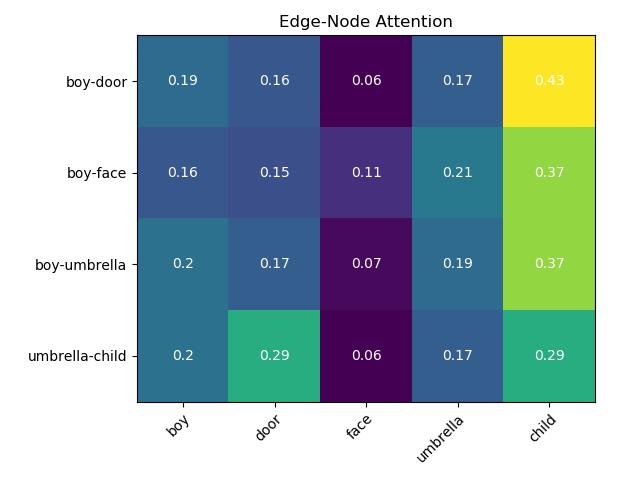}
        \hfill
        \includegraphics[width=0.45\linewidth]{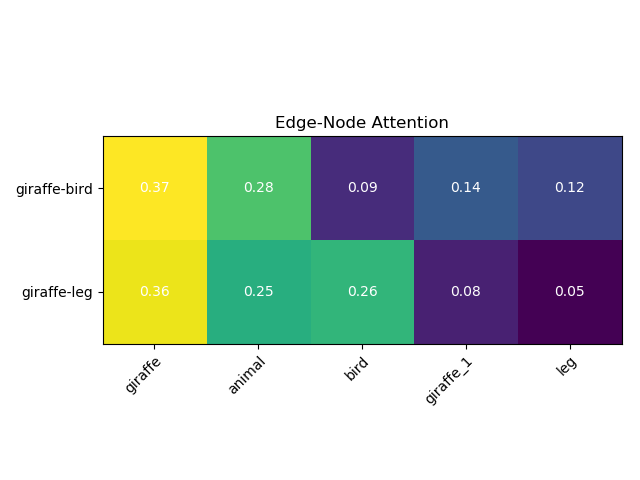}
        
        \caption{Edge to Node Attention heatmap}
     \end{subfigure}


    \vskip\baselineskip
     \begin{subfigure}[b]{\linewidth}
        \centering
        \includegraphics[width=0.45\linewidth, height=3cm]{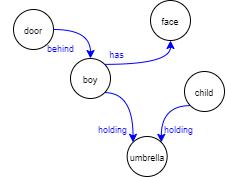}
        \hfill
        \includegraphics[width=0.15\linewidth, height=3cm]{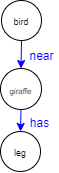}
        
        \caption{ Generated scene graphs}
     \end{subfigure}

\caption{Some positive example outputs from our network with associated attention map and scene graph.}
\label{fig:attention map positive}
\end{figure}

\begin{figure}[h]
\centering
    \begin{subfigure}[b]{\linewidth}
        \centering
        \includegraphics[width=0.45\linewidth, height=4cm]{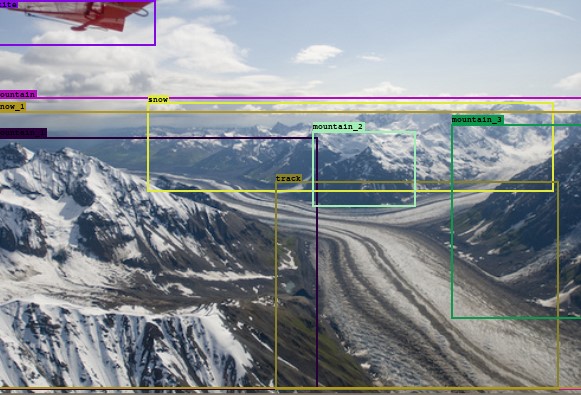}
        \hfill
        \includegraphics[width=0.45\linewidth, height=4cm]{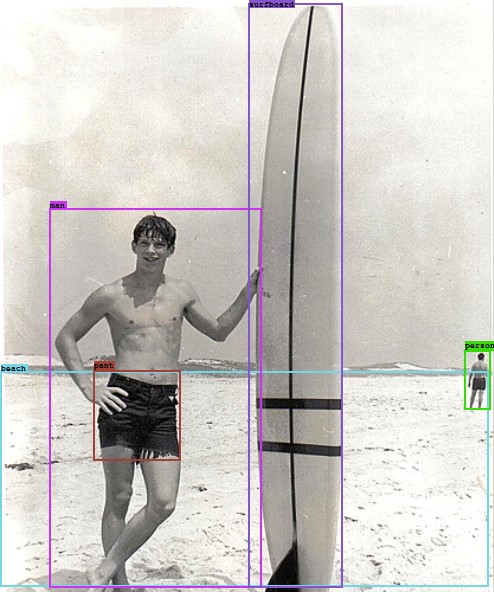}
        \caption{Scenes with objects and bounding boxes with respective labels}
    \end{subfigure}    
     \vskip\baselineskip
     \begin{subfigure}[b]{\linewidth}
        \centering
        \includegraphics[width=0.45\linewidth, keepaspectratio]{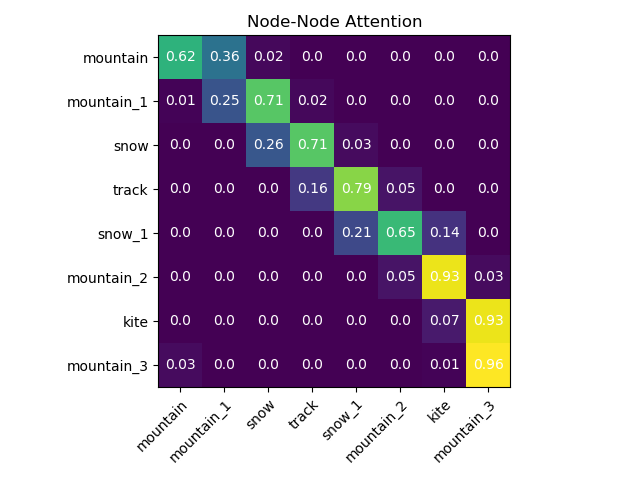}
        \hfill
        \includegraphics[width=0.45\linewidth, keepaspectratio]{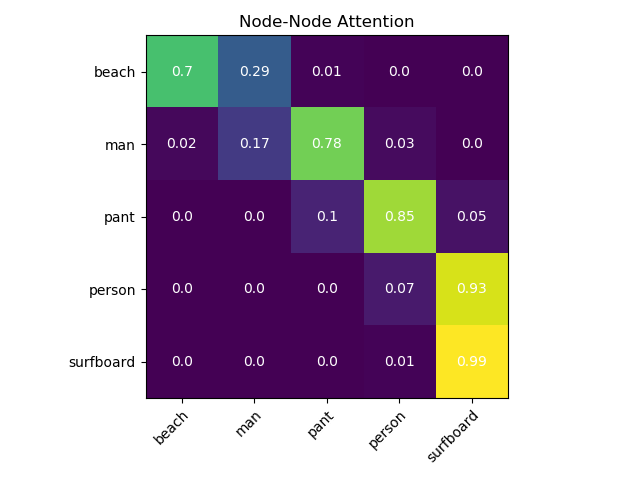}
        
        \caption{Node to Node Attention heatmap}
     \end{subfigure}
     
   \vskip\baselineskip
     \begin{subfigure}[b]{\linewidth}
        \centering
        \includegraphics[width=0.45\linewidth]{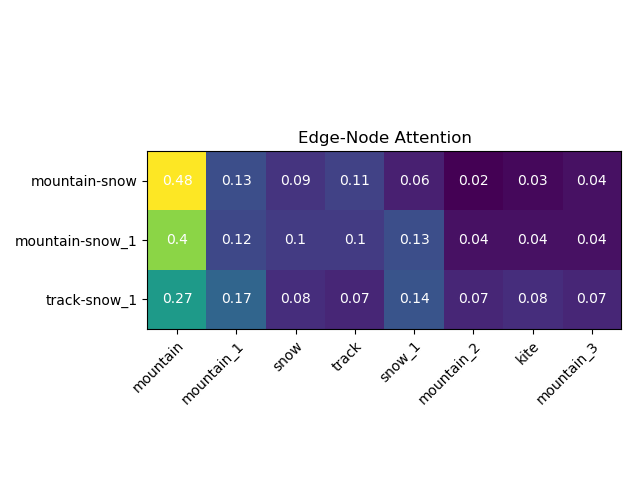}
        \hfill
        \includegraphics[width=0.45\linewidth]{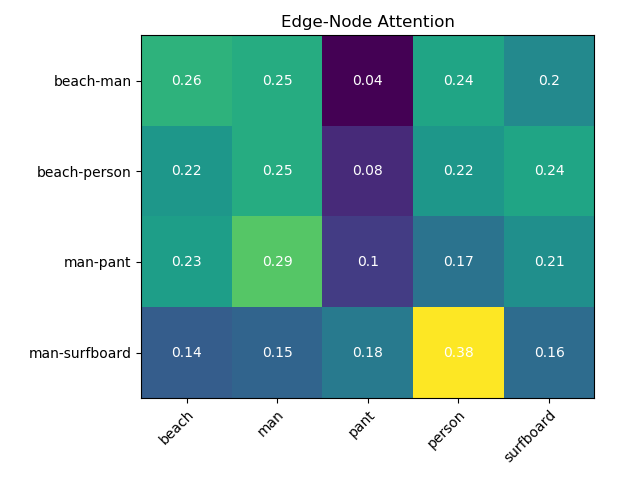}
        
        \caption{Edge to Node Attention heatmap}
     \end{subfigure}


   \vskip\baselineskip
    \begin{subfigure}[b]{\linewidth}
        \centering
        \includegraphics[width=0.45\linewidth]{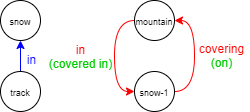}
        \hfill
        \includegraphics[width=0.45\linewidth]{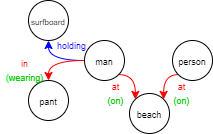}
        
        \caption{Generated scene graph. Here, blue ones are correctly predicted, red ones are mispredicted and green ones are the correct ground truth label for each mispredicted label. }
     \end{subfigure}

\caption{Some negative example outputs from our network with associated attention map and scene graph.}
\label{fig:attention map negative}
\end{figure}

\end{itemize}

\end{document}